\documentclass{ecai} 

\usepackage{latexsym}
\usepackage{amssymb}
\usepackage{amsmath}
\usepackage{amsthm}
\usepackage{booktabs}
\usepackage{enumitem}
\usepackage{graphicx}
\usepackage{color}

\usepackage[table]{xcolor}
\usepackage{arydshln}

\newcommand{\hypx}[0]{H$_{\text{px}}$}
\renewcommand{\paragraph}[1]{\noindent\textbf{#1}}

\begin{document}


\begin{frontmatter}



\title{HyenaPixel: Global Image Context with Convolutions}


\author[A]{\fnms{Julian}~\snm{Spravil}\thanks{Corresponding Author. Email: julian.spravil\,@\,iais.fraunhofer.de}}
\author[B,A]{\fnms{Sebastian}~\snm{Houben}}
\author[C,D,A]{\fnms{Sven}~\snm{Behnke}} 

\address[A]{Fraunhofer IAIS, Germany}
\address[B]{University of Applied Sciences Bonn-Rhein-Sieg, Germany}
\address[C]{University of Bonn, Computer Science Institute VI, Center for Robotics, Germany}
\address[D]{Lamarr Institute for Machine Learning and Artificial Intelligence, Germany}


\begin{abstract}
In computer vision, a larger effective receptive field (ERF) is associated with better performance.
While attention natively supports global context, its quadratic complexity limits its applicability to tasks that benefit from high-resolution input.
In this work, we extend Hyena, a convolution-based attention replacement, from causal sequences to bidirectional data and two-dimensional image space. 
We scale Hyena's convolution kernels beyond the feature map size, up to 191$\times$191, to maximize ERF while maintaining sub-quadratic complexity in the number of pixels.
We integrate our two-dimensional Hyena, HyenaPixel, and bidirectional Hyena into the MetaFormer framework.
For image categorization, HyenaPixel and bidirectional Hyena achieve a competitive ImageNet-1k top-1 accuracy of 84.9\% and 85.2\%, respectively, with no additional training data, while outperforming other convolutional and large-kernel networks.
Combining HyenaPixel with attention further improves accuracy.
We attribute the success of bidirectional Hyena to learning the data-dependent geometric arrangement of pixels without a fixed neighborhood definition.
Experimental results on downstream tasks suggest that HyenaPixel with large filters and a fixed neighborhood leads to better localization performance.
\end{abstract}

\end{frontmatter}

\section{Introduction}
\label{sec:intro}

The 35-year history of Convolutional Neural Networks' (ConvNets)~\cite{lecun1989backpropagation} successful track record~\cite{lecun1998gradient,Behnke03,CiresanMS12,krizhevsky2012imagenet,szegedy2015going,he2016deep,tan2019efficientNet} has recently been challenged by Vision Transformers (ViTs)~\cite{dosovitskiy2020image}.
The ViT plays a significant role in the recent improvements in computer vision~\cite{tu2022maxvit,woo2023convnext,zhu2023biformer} due to its simple architecture: The input image is split into equal-sized patches further processed by a regular transformer encoder with bidirectional attention~\cite{vaswani2017attention}.
This design scales well in terms of data and parameters, shares a similar architecture across modalities, and achieves remarkable performance in a self-supervised setting.
Under the pressure of competition, ConvNets are currently reassessed.
For example, new evidence suggests that ConvNets follow similar scaling laws~\cite{smith2023convnets,woo2023convnext,wang2023internimage}.
On the other hand, convolution serves as a source of inspiration for ViT enhancements.
For instance, the adoption of the hierarchical network layout led to significant improvements~\cite{liu2021swin}.
Hybrid models emerged that apply convolution in earlier layers~\cite{yu2022metaformer_baselines} or as a replacement of or addition to the Feed Forward Network (FFN) in each transformer block~\cite{tu2022maxvit, zhu2023biformer}.
Other improvements focus on mimicking properties of convolution with attention.
This includes attention on local windows~\cite{liu2021swin} or sparse grids~\cite{tu2022maxvit}.
Furthermore, attention can be replaced with computationally cheaper alternatives.
These replacements focus, among others, on the Fourier transform~\cite{lee2021fnet}, simple pooling~\cite{yu2022metaformer} or local convolutions~\cite{yu2022metaformer_baselines}.

Token mixers with sub-quadratic complexity are highly sought after, as image resolution is one of the most important performance factors for image classification~\cite{tu2022maxvit}, vision language modeling~\cite{mckinzie2024mm1}, and other downstream tasks.
Currently, attention requires specialized strategies, such as subdividing input images followed by separate processing~\cite{lin2023sphinx, mckinzie2024mm1}, potentially limiting image context.
Alternatively, to aggregate information over the entire input with small efficient local operations, a deep network is essential~\cite{szegedy2015going,he2016deep,tan2019efficientNet}.
A promising new path is the integration of large convolutional filters for sequence modeling~\cite{poli2023hyena,fu2023simple} and also for vision with medium~\cite{peng2017large,liu2022convnet,guo2022segnext} to large kernels sizes---up to 61$\times$ 61~\cite{ding2022scaling,liu2023more}.

\begin{figure}[t]
\centering
\includegraphics[width=.95\linewidth]{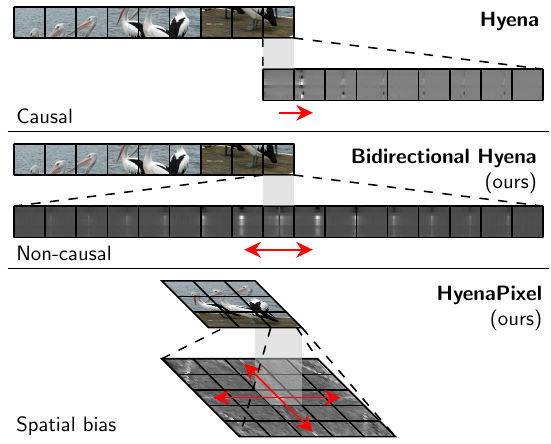}
\caption{%
Our extensions of Hyena~\protect\cite{poli2023hyena} (top).
In bidirectional Hyena (center), a large non-causal filter is applied to both sides of the token sequence.
HyenaPixel (bottom) uses a large convolutional kernel to process 2D feature maps.
We show the evaluation of the rightmost token position and the resulting kernel overlap.}
\label{fig:preview}
\end{figure}

In this work, we explore the Hyena operator~\cite{poli2023hyena} as an attention replacement in vision applications.
The Hyena operator uses long convolutions with gating and was originally proposed for causal language modeling.
This token mixer qualifies for this exploration because of its sub-quadratic complexity with respect to input sequence length and its use of convolution, native to computer vision.
In addition, Hyena has a similar intuition as attention: It provides global context by computing a weighted sum, data-driven for attention and learned for Hyena, over all input tokens for each output token.
In this setting, we ask two research questions:
i) Is an approximation of attention with fixed learned attention patterns, like the Hyena operator, a sufficient replacement for fine-granular, fully data-driven attention in vision applications? ii) Does the addition of a fixed pixel neighborhood or spatial bias impact performance?

Fig.~\ref{fig:preview} illustrates our approach.
We extend the causal convolution-based attention replacement Hyena~\cite{poli2023hyena} by considering bidirectional, non-causal information flow---bidirectional Hyena (H$_\text{b}$)---and by accommodating the 2D nature of images with spatial bias---HyenaPixel (\hypx{}). 

\noindent The main contributions of our work are:
\begin{itemize}
\item We extend causal convolution-based Hyena~\cite{poli2023hyena} to non-causal and 2D inputs, while maintaining training stability, sub-quadratic complexity, and enabling large effective receptive fields (ERFs). 
\item We evaluate the resulting token mixers H$_\text{b}$ and \hypx{} in the MetaFormer framework for image classification, object detection, and semantic segmentation and achieve results outperforming other large-kernel networks.
\item We analyze the learned features of \hypx{}, elaborate on the importance of global context, bidirectional modeling and spatial bias with convolution and compare our approach with different token mixer configurations.
\end{itemize}

\section{Related Work}
\label{sec:related_work}

Improvements to the ViT~\cite{dosovitskiy2020image} focus on the architecture~\cite{liu2021swin,yu2022metaformer}, training strategy~\cite{woo2023convnext}, and attention mechanism or generally token mixing~\cite{liu2021swin,dong2022cswin,tu2022maxvit,yu2022metaformer,yu2022metaformer_baselines,zhu2023biformer}.
By following a four-stage architecture with convolution-based down-sampling layers, the hierarchical structure provides a consistent accuracy improvement~\cite{liu2021swin}.
Knowledge distillation with an extra teacher token from a CNN teacher also proved helpful~\cite{touvron2021training}.
There are different variants of attention for visual data:
Swin Transformers~\cite{liu2021swin} apply attention to shifted rectangular windows while MaxViT~\cite{tu2022maxvit} uses window attention and sparse grid attention for global interactions.
CSWin~\cite{dong2022cswin} uses parallel row and column attention with integrated position enhancement.
BiFormer~\cite{zhu2023biformer} implements data-driven key-value filtering to reduce computational overhead for irrelevant tokens.
Similarly, DAT~\cite{xia2022vision} selects important tokens based on fixed reference points and predicted offsets.
Some methods use convolutional layers within each transformer block to enhance local positional information~\cite{dong2022cswin, xia2022vision, zhu2023biformer} while others replace the FFN with a convolutional component~\cite{tu2022maxvit}.
Focus of current research is the self-supervised learning of visual features~\cite{woo2023convnext}.

\paragraph{ConvNets and large kernels.}
ConvNets first proposed in the 1980s~\cite{lecun1989backpropagation} are responsible for may advancements in computer vision~\cite{lecun1998gradient,Behnke03,CiresanMS12,krizhevsky2012imagenet,szegedy2015going,he2016deep,tan2019efficientNet}.
New progress has been made in ConvNet research as a result of the new success of the transformers~\cite{vaswani2017attention, dosovitskiy2020image}.
Typically, the transformer architecture is used as a basis while the attention layer is replaced with a combination of convolutional layers~\cite{yu2022metaformer,yu2022metaformer_baselines,wang2023internimage}.
For instance, InternImage~\cite{wang2023internimage} replaces attention with deformable convolutions to realize long range data-driven dependencies and scale the model to one billion parameters.
ConvNeXt~\cite{liu2022convnet} builds on a deep stack of small convolutional blocks, that later proved suitable for unsupervised training as masked auto encoder~\cite{woo2023convnext}.

More recent research investigates ConvNets with large kernels.
Common across these networks is their regularization through parameterizing the convolution weights to guarantee smoothness~\cite{romero2021flexconv,romero2021ckconv,fu2023simple,poli2023hyena} or by applying sparsity of some form~\cite{dai2017deformable,peng2017large,guo2022segnext,liu2023more}.
\citet{romero2021flexconv} proposed parameterized filters with dynamic size and discovered that the filter size increases with depth.
However, parameterized kernels as used in \cite{romero2021flexconv,romero2021ckconv,fu2023simple,poli2023hyena} require assumptions about how the input is processed.
The global convolution network~\cite{peng2017large} applies separable convolutions (21$\times$1 and 1$\times$21) to improve classification while maintaining localization for semantic segmentation.
SegNeXt~\cite{guo2022segnext} also utilizes parallel separable convolutions with sizes between $7$ and $21$.
RepLKNet~\cite{ding2022scaling} uses full convolutions with size up to 31$\times$31, while the large kernels are fused by re-parameterization of multiple smaller kernels.
SLaK~\cite{liu2023more} proposes two parallel kernels spanning 61$\times$5 with dynamic sparsity.
However, dynamic sparsity, which theoretically reduces the multiply-accumulate operations (MACs), requires an efficient hardware implementation, still being sought.

\paragraph{Substitutes for attention.}
While attention is a powerful and flexible mechanism, its complexity is quadratic in the number of tokens~\cite{vaswani2017attention}.
Linear attention~\cite{katharopoulos2020transformers} uses a kernel formulation to express similarity between tokens.
However, finding expressive kernel functions is challenging~\cite{han2023flatten}.
MLP-Mixer~\cite{tolstikhin2021mixer} uses multiple linear layer stacks applied alternating on the channel and token dimensions.
The idea of basic token mixing is further extended to a mean-pooling approach~\cite{yu2022metaformer} and simple convolutional layers~\cite{yu2022metaformer_baselines}.
FNet~\cite{lee2021fnet} replaces the attention layer with the Fourier transform along the token and channel dimensions.
Hyena~\cite{poli2023hyena} uses long and short convolutions for causal token mixing.
They share the same goal as \citet{fu2023simple} to apply convolutions for efficient training with long token sequences.
Convolution appears to be a promising solution for vision-related~\cite{yu2022metaformer_baselines} but also sequence-modeling~\cite{poli2023hyena} tasks as many other alternatives struggle to achieve high performance.

The simultaneous work by \citet{zimerman2023multi}, like ours, aims to raise the dimensional extent of Hyena.
The authors evaluation on small-scale datasets in different transformer frameworks.
Their approach improved the performance over their baselines, but also benefited from additional subsequent attention layers.
The causality of Hyena is addressed by rotating the input after each layer.
We propose a non-causal Hyena layer that does not require input transformations like rotation and can be also applied to higher-dimensional input.
While the authors showed improved classification performance for small datasets by adding spatial bias, we find that this is not the case for larger corpora.
In this case, we show that sequential bidirectional data modeling is superior.

\begin{figure}[t]
\centering
\includegraphics[width=0.95\linewidth]{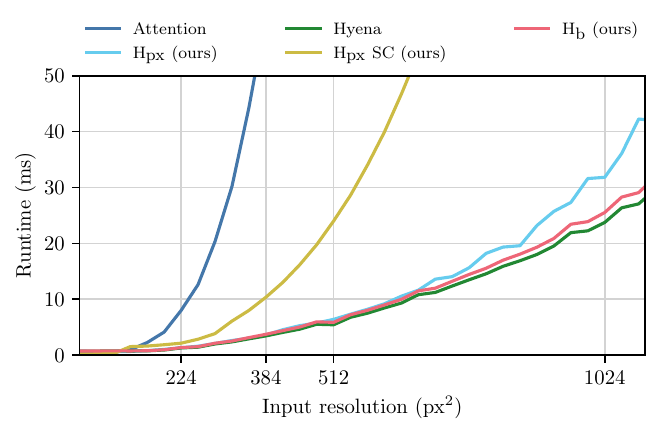}
\vspace{-1em}
\caption{Runtime  scaling of token mixers with global token interactions. 
Input images are patched with a patch size of 4.
\hypx{} SC uses separable convolutions (SC) instead of an implicit filter.
The experiment was conducted on an Nvidia A100 GPU.
}
\label{fig:runtime}
\end{figure}

\begin{figure*}[t]
\centering
\includegraphics[width=0.95\linewidth]{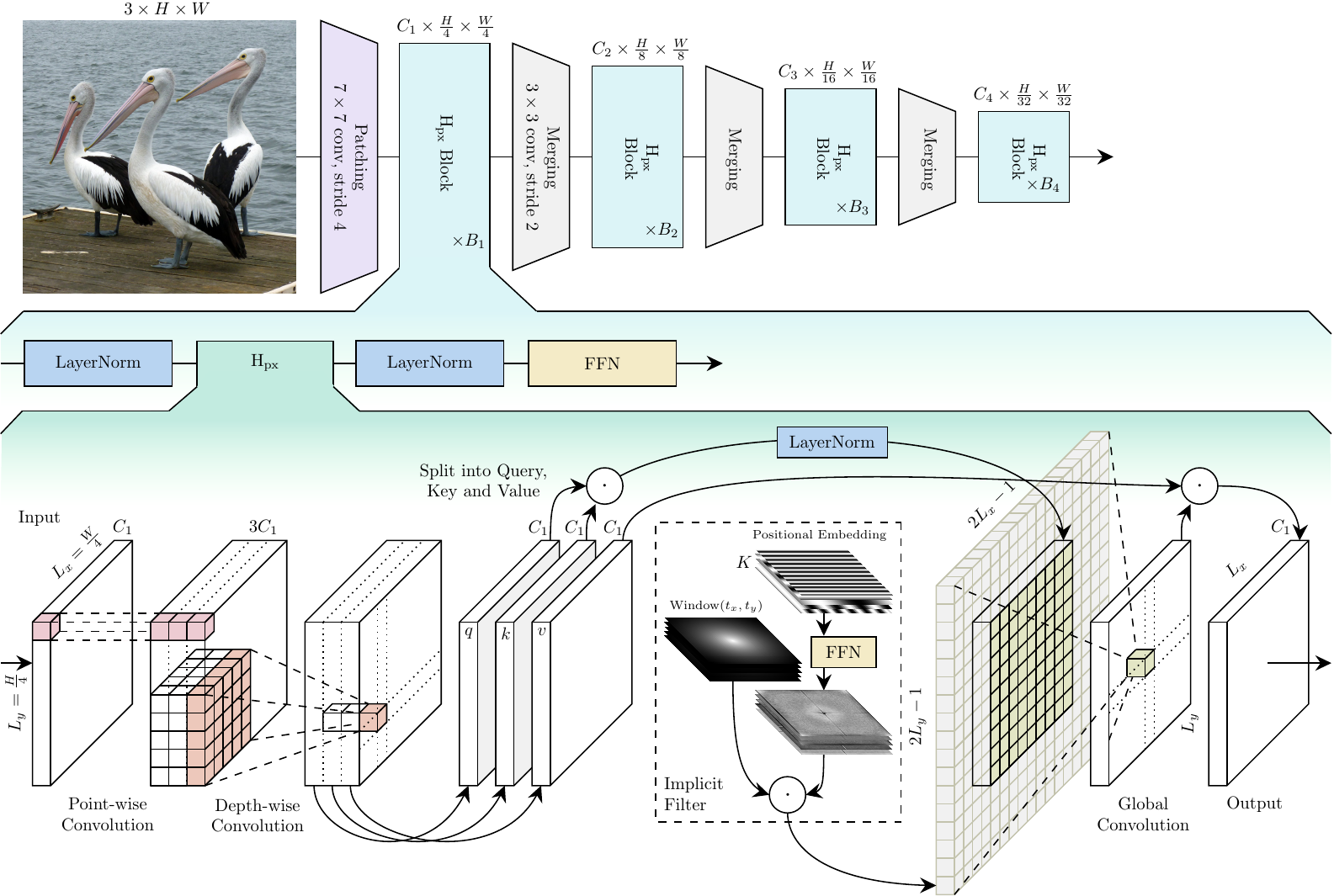}
\caption{\textbf{The HyenaPixel (\hypx{}) operator embedded in the MetaFormer framework.}
The first row shows the MetaFormer framework~\protect\cite{yu2022metaformer_baselines} with an input of size $H\!\times\!W$, typically set to $224\,\text{px}^2$.
The input is divided into 4$\times$4 patches and processed by a sequence of \hypx{} blocks with intermediate merging layers to reduce spatial resolution.
The second row focuses on the structure of the \hypx{} block with Layer Norm~\protect\cite{ba2016layer} and a Feed Forward Network (FFN).
The last row shows the \hypx{} operator.
The input feature map has two spatial dimensions $L_y = H / 4$ and $L_x= W / 4$ and the channel dimension $C_1$. 
First, the dimension is increased to $3C_1$ by a point-wise and a depth-wise 5$\times$5 convolution. 
The resulting feature map is split into three equal-sized chunks: query $q$, key $k$, and value $v$. 
The result of the element-wise multiplication $\odot$ of $q$ and $k$ is normalized and convolved with a global implicit filter.
The final output is the element-wise multiplication with $v$.
}
\label{fig:hyena_pixel_operator}
\end{figure*}

\section{Method}
\label{sec:method}

\paragraph{Motivation.}
Vision Transformers (ViTs) are one point ahead of Convolutional Neural Networks (ConvNets): 
A single attention layer already has global context.
Current ConvNets scale kernels to at most 61$\times$61~\cite{ding2022scaling,liu2023more} and thus only give the center pixel full context.
Kernels that are larger than the feature map, however, proved beneficial~\cite{ding2022scaling}.
A recent approach designed for language modeling promises global context based on gated global convolution, namely the Hyena operator~\cite{poli2023hyena}.
Motivated by Hyena's promising properties for sequence modeling, we apply it to the 2D pixel space with drastically larger kernels than previously considered.

\paragraph{Hyena.}
The Hyena operator by \citet{poli2023hyena} first projects the input sequence $x$ of length $L$ into different spaces $p_0(x), \dots, p_{O}(x)$.
The number of projections is determined by the order parameter $O$.
The projection $p_i(\cdot)$ is defined by a linear layer and a local convolution.
Aggregation of the output is handled recursively by element-wise multiplication of the previous result with the next projection:
\begin{equation}
   y_{i + 1} = g(y_{i}) \cdot p_{i + 2}(x).
\end{equation}
Intermediate results are convolved with a large implicit filter, handled by function $g(\cdot)$.
The initial value is set to $y_0 = p_0(x) \cdot p_1(x)$.
The kernel weights of the global convolution are implicitly modeled by applying a FFN with a sinusoidal activation function to a positional embedding.
The positional embedding is a truncated complex exponential basis $\rho_k(t) = e^{i2 \pi kt/L}$ for $k = 0, \ldots, K - 1$, where $K$ represents the embedding dimension.
Furthermore, the resulting filter is regularized by modulation with an exponential decay
\begin{equation}\label{eq:window_1d}
   \text{Window}(t) = \exp \{ - \alpha t \} + b \text{,}
\end{equation}
with scaling factor $\alpha$, bias $b$ and $t = 0, 1, \ldots, L - 1$.
Causality is achieved by circularizing the filter by zero-padding to length $2L-1$ and keeping the $L$ left output positions of the circular FFT-based convolution.
In this work, we simplify the Hyena operator by setting $O=2$.
With this simplification, we can rewrite the recursive formulation as follows:
\begin{equation}
   y = g(q \cdot k) \cdot v \text{,}
\end{equation}
with query $q=p_0(x)$, key $k=p_1(x)$ and value $v=p_2(x)$.

\paragraph{Bi-directional Hyena (H$_{\text{b}}$).}
The concept of causality, i.e. the next token in a sequence can only refer to the previous tokens, is very useful for autoregressive language modeling but unnatural for offline-processing of signals.
Therefore, we extend the Hyena operator to bidirectional sequence modeling.
This requires larger filters and an evaluation region centered around the current sequence element.
In detail, the filter width is increased from sequence length $L$ to $2L-1$ to have full sequence coverage at each token position.
Hyena pads the filter and input with zeros to $2L-1$, which, in our case, is only required for the input.
Instead of selecting the output indices $0, 1, \ldots, L - 1$ of the circular FFT-based convolution, we select the indices $\frac{L}{2}, \frac{L}{2} + 1, \ldots, L + \frac{L}{2}$ to obtain a centered filter.
The modulation of the filter $\text{Window}(t)$ can be set to $\exp \{ - \alpha |t| \} + b$ with $t = - L + 1, - L + 2, \ldots,  0, \ldots, L - 1$.
Note that the complexity is the same as for the causal Hyena operator, i.e. $\mathcal{O}(L\log_2{}L)$.
We name the resulting operator H$_{\text{b}}$.

\paragraph{HyenaPixel (\hypx{}).}
Images are two-dimensional and could therefore benefit from a fixed pixel neighborhood.
To add spatial bias to H$_{\text{b}}$, we replace $\text{Window}(t)$ (cf. Equation~\ref{eq:window_1d}) with
\begin{equation}
\resizebox{.91\linewidth}{!}{$
\text{Window}(t_x, t_y)  = \exp \left\{ - \alpha \sqrt{ \left(t_x - c_x\right)^2 + \left(t_y - c_y\right)^2 }\right\} + b \text{.}
$}
\end{equation}
where $c_x$ and $c_y$ is the filter center.
We use the 2D extension of the 1D positional encoding of the original Transformer proposed by \citet{wang2021translating}.
Positions are encoded in the vertical and horizontal direction using sine and cosine functions.
The inputs to the circular 2D FFT convolution are zero padded to $(2L_x-1) \times (2L_y-1)$.
The asymptotic complexity of \hypx{} is $\mathcal{O}( L_x L_y \log_2{} \left(L_x L_y) \right)$.
In practice, however, \hypx{} is slightly slower while the performance only differs marginally for resolutions below $512\,\text{px}^2$, cf. Fig.~\ref{fig:runtime}.
We name this extension HyenaPixel (\hypx{}).

\paragraph{Hierarchical transformer.}
We embed H$_\text{b}$ and \hypx{} in a Transformer encoder~\cite{vaswani2017attention,dosovitskiy2020image}, considering four frameworks: the original isomorphic ViT~\cite{dosovitskiy2020image} and three hierarchical models MetaFormer~\cite{yu2022metaformer,yu2022metaformer_baselines}, ConvNeXt~\cite{liu2022convnet,woo2023convnext}, and Swin Transformer~\cite{liu2021swin}.
However, because hierarchical models consistently perform better than their isomorphic counterparts~\cite{liu2022convnet} and MetaFormer already explored different token mixer types, we settled on the MetaFormer architecture, as depicted in Fig.~\ref{fig:hyena_pixel_operator}.

There are a few key differences to the Swin Transformer:
First, the image patching layer and the in-between patch merging layers have an overlap (i.e. the kernel size is larger than the stride).
Second, the depth of the network is increased while the width is decreased.
Finally, the commonly used activation function GELU~\cite{hendrycks2016gaussian} is replaced with StarReLU~\cite{yu2022metaformer_baselines}.

\paragraph{Model sizes.}
We explore the following model sizes:
\vspace{-0.3em}
\begin{itemize}
\item S4: $C = (64, 128, 320, 512)$, $B = (1, 1, 1, 1)$;
\item S12: $C = (64, 128, 320, 512)$, $B = (2, 2, 6, 2)$;
\item S18: $C = (64, 128, 320, 512)$, $B = (3, 3, 9, 3)$; and
\item B36: $C = (128, 256, 512, 768)$, $B = (3, 12, 18, 3)$.
\end{itemize}
\vspace{-0.3em}
Here, $C$ is the channel dimension and $B$ is the number of blocks per stage.
We use the syntax of \citet{yu2022metaformer_baselines} and classify the channel dimensionality with the letter S (small) followed by the total number of blocks $ \lVert B \rVert_1$.
The full model is depicted in Fig.~\ref{fig:hyena_pixel_operator}.

\paragraph{Token mixer layout.}
The main layout has H$_\text{b}$ or \hypx{} in each stage of the network, that is H$_\text{b}$Former and \hypx{}Former.
The hyper-parameters of H$_\text{b}$  are set to filter sizes $L=\left[2 \cdot 56^2- 1, 2 \cdot 28^2- 1, 2 \cdot 14^2 - 1, 2 \cdot 7^2 - 1\right]$, position embedding dimensions $K=\left[32, 32, 48, 64\right]$, and hidden filter projection dimensions of $2K_i$ for each stage $i$.
\hypx{} parameters are similar, with the difference that the kernel size is defined by $L_x=L_y=\left[111, 55, 27, 13\right]$.
Global context, as provided by attention, proved beneficial in later stages~\cite{yu2022metaformer_baselines,dong2022cswin}.
Inspired by this observation, we also formulate the C\hypx{}Former, with local convolutions in the first two and \hypx{} in the last two stages.
The local convolution follows the inverse separable convolution proposed in MobileNetV2~\cite{sandler2018mobilenetv2} that is also employed in the ConvFormer~\cite{yu2022metaformer} with a kernel size of 7.
Furthermore, we propose \hypx{}AFormer to evaluate whether attention has any additional value beyond the capabilities of \hypx{}.
\hypx{}AFormer uses \hypx{} in the first two stages, followed by the attention stages.
Similarly, we define H$_\text{b}$AFormer.

\section{Evaluation}
\label{sec:exp}

\subsection{Image Classification}
\label{sec:eval:image_classifciation}

\paragraph{Training on ImageNet-1k.}
We train on ImageNet-1k~\cite{deng2009imagenet} (IN-1k) consisting of 1.3M and 50K images in the training and validation set, respectively.
The images are categorized into 1000 classes.
We follow the training strategy of \citet{yu2022metaformer_baselines} and optimize with AdamW~\cite{loshchilov2017decoupled}, a batch size of 4096, a learning rate of $4e^{-3}$, and a weight decay of $0.05$ for 310 epochs.
The learning rate is scheduled with a linear warm-up for 20 epochs followed by a cosine decay for 280 epochs and an additional 10 cool-down epochs with a final learning rate of $1e^{-5}$.
Regularization is added by stochastic depth~\cite{huang2016deep} (0.6 for B36 scale, otherwise 0.2), label smoothing~\cite{szegedy2016rethinking} with 0.1, and res scale~\cite{shleifer2021normformer} in the last two stages.
We do not apply token labeling~\cite{jiang2021all}.
We apply the following data augmentations: Mixup~\cite{zhang2017mixup}, Cutmix~\cite{yun2019cutmix}, Rand\-Augment~\cite{cubuk2020randaugment}, and Random Erasing~\cite{zhong2020random}.
Our implementation is based on the \textit{timm} framework~\cite{rw2019timm}. 

\paragraph{Fine-tuning on higher resolution.}
ConvNets naturally scale to different resolutions and can show improved accuracy for higher resolution inputs~\cite{tan2019efficientNet}.
This also applies to \hypx{}Former.
On the other hand, H$_\text{b}$Former specializes in the specific input shape and would require an interpolation of the learned one-dimensional filters.
Note that a similar procedure is required for ViTs where the positional embedding needs to be resampled~\cite{dosovitskiy2020image}.

We fine-tune \hypx{}Former-S18 on IN-1k with the resolutions $384\,\text{px}^2$ and $512\,\text{px}^2$.
Resampling the filters of \hypx{}Former-S18 to the sizes $L_x=L_y=\left[191, 95, 47, 23\right]$ showed no significant improvement for $384\,\text{px}^2$, therefore, we focus on the direct approach.
In accordance with the standard procedure~\cite{yu2022metaformer_baselines}, we fine-tune for 30 epochs with AdamW, a learning rate of $5e-5$, a batch size of 1024, exponential moving average~\cite{polyak1992acceleration} and head dropout of 0.4.
Learning rate scheduling, Mixup, and Cutmix are disabled.

\begin{table}[!ht]
\caption{
\textbf{IN-1k validation set results with input resolutions of $224\,\text{px}^2$.}
We compare different attention (A), convolution (C), and hybrid (H) approaches.
The approaches are categorized into the following groups based on the computational requirements: up to 8G MACs, 8-12G MACs, 12-18G MACs, and more than 18G MACs.
MACs are calculated using \texttt{fvcore}~\protect\cite{fvcore}.
The entries in each group are sorted in ascending order by the primary key ``Top-1 accuracy'' and in descending order by the secondary key ``MACs''.
Note that the reported parameter count and MACs of SLaK~\cite{liu2023more} marked with a ``*'' require specialized hardware supporting sparse convolution.
Our models are highlighted in \colorbox{gray!20}{gray}.
}
\label{tab:imagenet}
\centering
\addtolength{\tabcolsep}{-3.pt}
\begin{tabular}{llrrr}
\toprule[1pt]
   Model & Type & \#Param. & MACs & Top-1 \\
\midrule[1pt] 

Swin-T~\cite{liu2021swin}  & A & 28M & 4.5G & 81.4 \\
DAT-T~\cite{xia2022vision} & A & 29M & 4.6G & 82.0 \\
CSWin-T~\cite{dong2022cswin} & A & 23M & 4.3G & 82.7 \\
CSWin-S~\cite{dong2022cswin} & A & 35M & 6.9G & 83.6 \\

\noalign{\vskip 0.5ex}\hdashline\noalign{\vskip 0.5ex}
ConvNeXt-T~\cite{liu2022convnet} & C & 29M & 4.5G & 82.1 \\
SLaK-T~\cite{liu2023more} & C & *30M & *5.0G & 82.5 \\ 
\rowcolor{gray!20!}
C\hypx{}Former-S18 & C & 28M & 4.3G & 83.0 \\ 
ConvFormer-S18~\cite{yu2022metaformer_baselines} & C & 27M & 3.9G & 83.0 \\
\rowcolor{gray!20!}
\hypx{}Former-S18 & C & 29M & 4.9G & 83.2  \\
InternImage-T~\cite{wang2023internimage} & C & 30M & 5.0G & 83.5 \\ 
\rowcolor{gray!20!}
H$_{\text{b}}$Former-S18 & C & 28M & 4.4G & 83.5 \\

\noalign{\vskip 0.5ex}\hdashline\noalign{\vskip 0.5ex}
\rowcolor{gray!20!}
H$_{\text{b}}$AFormer-S18 & H & 27M & 4.4G & 83.2 \\ 
MaxViT-T~\cite{tu2022maxvit} & H & 31M & 5.6G & 83.6 \\
\rowcolor{gray!20!}
\hypx{}AFormer-S18 & H & 28M & 4.7G & 83.6 \\ 
CAFormer-S18~\cite{yu2022metaformer_baselines} & H & 26M & 4.1G & 83.6 \\ 
BiFormer-S~\cite{zhu2023biformer} & H & 26M & 4.5G & 83.8 \\ 

\midrule

Swin-S~\cite{liu2021swin}  & A & 50M & 8.7G & 83.3 \\ 
DAT-S~\cite{xia2022vision} & A & 50M & 9.0G & 83.7 \\

\noalign{\vskip 0.5ex}\hdashline\noalign{\vskip 0.5ex}
ConvNeXt-S~\cite{liu2022convnet} & C & 50M & 8.7G & 83.2 \\ 
SLaK-S~\cite{liu2023more} & C & *55M & *9.8G & 83.8 \\
\rowcolor{gray!20!}
\hypx{} / Conv & C & 56M & 8.8G & 84.0 \\
ConvFormer-S36~\cite{yu2022metaformer_baselines} & C & 40M & 7.6G & 84.0 \\
InternImage-S~\cite{wang2023internimage} & C & 50M & 8.0G & 84.2 \\

\noalign{\vskip 0.5ex}\hdashline\noalign{\vskip 0.5ex}
BiFormer-B~\cite{zhu2023biformer} & H & 59M & 9.8G & 84.3 \\
MaxViT-S~\cite{tu2022maxvit} & H & 69M  & 11.7G & 84.5 \\
CAFormer-S36~\cite{yu2022metaformer_baselines} & H & 39M & 8.0G & 84.5 \\

\midrule

Swin-B~\cite{liu2021swin}  & A & 88M & 15.4G & 83.6 \\
DAT-B~\cite{xia2022vision} & A & 88M & 15.8G & 84.0 \\
CSWin-B~\cite{dong2022cswin} & A & 78M & 15.0G & 84.2 \\

\noalign{\vskip 0.5ex}\hdashline\noalign{\vskip 0.5ex}
RepLKNet-31B~\cite{ding2022scaling} & C & 79M & 15.3G & 83.5 \\ 
ConvNeXt-B~\cite{liu2022convnet} & C & 89M & 15.4G & 83.9 \\
SLaK-B~\cite{liu2023more} & C & *95M & *17.1G & 84.0 \\
ConvFormer-M36~\cite{yu2022metaformer_baselines} & C & 57M & 12.8G & 84.5 \\

\noalign{\vskip 0.5ex}\hdashline\noalign{\vskip 0.5ex}
\rowcolor{gray!20!}
\hypx{} / Conv / H$_{\text{b}}$ / CA & H & 111M & 17.3G & 84.7 \\
CAFormer-M36~\cite{yu2022metaformer_baselines} & H & 56M & 13.2G & 85.1 \\

\midrule

ConvNeXt-L~\cite{liu2022convnet} & C & 198M & 34.4G & 84.3 \\
ConvFormer-B36~\cite{yu2022metaformer_baselines} & C & 100M & 22.6G & 84.8 \\
InternImage-B~\cite{wang2023internimage} & C & 97M & 18.0G & 84.9 \\
\rowcolor{gray!20!}
\hypx{}Former-B36 & C & 111M & 25.3G & 84.9 \\
\rowcolor{gray!20!}
H$_{\text{b}}$Former-B36 & C & 102M & 23.8G & 85.2 \\

\noalign{\vskip 0.5ex}\hdashline\noalign{\vskip 0.5ex}
MaxViT-B~\cite{tu2022maxvit} & H & 120M & 23.4G & 85.0 \\
MaxViT-L~\cite{tu2022maxvit} & H & 212M & 43.9G & 85.2 \\
CAFormer-B36~\cite{yu2022metaformer_baselines} & H & 99M & 23.2G & 85.5 \\

\bottomrule[1pt]
\end{tabular}
\end{table}

\begin{table}[ht]
\caption{
\textbf{IN-1k validation set results with input resolutions of $384\,\text{px}^2$ and $512\,\text{px}^2$.}
}
\label{tab:imagenet_hs}
\centering
\addtolength{\tabcolsep}{-3.pt}
\begin{tabular}{llrrr}
\toprule[1pt]
   Model & Type & \#Param. & MACs & Top-1 \\
\midrule[1pt]
\multicolumn{5}{l}{$384\,\text{px}^2$} \\

\midrule[1pt]

ConvFormer-S18~\cite{yu2022metaformer_baselines} & C & 27M & 11.6G & 84.4 \\
\rowcolor{gray!20!}
\hypx{}Former-S18 & C & 29M & 12.9G & 84.7 \\ 

\noalign{\vskip 0.5ex}\hdashline\noalign{\vskip 0.5ex}
CAFormer-S18~\cite{yu2022metaformer_baselines} & H & 13.4G & 85.0 \\
MaxViT-T~\cite{tu2022maxvit} & H & 31M & 17.7G & 85.2 \\

\midrule[1pt]
\multicolumn{5}{l}{$512\,\text{px}^2$} \\
\midrule[1pt]

\rowcolor{gray!20!}
\hypx{}Former-S18 & C & 29M & 22.3G & 84.8 \\ 
\noalign{\vskip 0.5ex}\hdashline\noalign{\vskip 0.5ex}
MaxViT-T~\cite{tu2022maxvit} & H & 31M & 33.7G & 85.7 \\

\bottomrule[1pt]
\end{tabular}
\end{table}

\paragraph{Results on ImageNet-1k.}
Tab.~\ref{tab:imagenet} reports the results on IN-1k for $224\,\text{px}^2$ images.
For validation, a center-cropped region of the input image is selected with a crop size between 0.8 and 1.0 that maximizes the accuracy.
Reference methods are selected based on a comparable training strategy and computational requirement.

We have three models that qualify as ConvNets: H$_\text{b}$Former, \hypx{}Former, and C\hypx{}Former.
Our best model, H$_\text{b}$Former, outperforms other strong ConvNets, namely ConvNeXt~\cite{liu2022convnet}, SLaK~\cite{liu2023more}, and ConvFormer~\cite{yu2022metaformer_baselines}, and achieves on par performance to InternImage~\cite{wang2023internimage} on small scale (InternImage-T, 5.0G MACs, $83.5\%$ accuracy) and even surpasses it by $0.3\%$ on a larger scale (InternImage-B, 18.0G MACs, $84.9\%$ accuracy) with an accuracy of $85.2\%$.
In comparison to attention-based and hybrid models, the H$_\text{b}$Former shows competitive performance.
On a small scale, the BiFormer-S~\cite{zhu2023biformer} surpasses the H$_\text{b}$Former-S18 by $0.3\%$, while it loses its advantage with increasing scale.
H$_\text{b}$Former-B36 is on par with the MaxViT-L~\cite{tu2022maxvit} (43.9G MACs, $85.2\%$ accuracy) while requiring $52\%$ and $46\%$ fewer parameters and MACs, respectively.
However, the CAFormer-B36~\cite{yu2022metaformer_baselines} is $0.3\%$ accuracy points ahead.

The addition of a fixed neighborhood definition to the token mixer slightly reduces the categorization performance.
By using 2D convolutions, we observe a drop of $0.3\%$ between H$_\text{b}$Former-B36 and \hypx{}Former-B36.

Combining \hypx{} or H$_\text{b}$ with attention following CAFormer~\cite{yu2022metaformer_baselines} leads to mixed results.
H$_\text{b}$ is incompatible with attention, leading to an $0.4\%$ advantage of CAFormer-S18 over H$_\text{b}$AFormer-S18.
We assume that the local positional information learned by the earlier H$_\text{b}$ layers is not representative enough.
On the other hand, replacing H$_\text{b}$ with \hypx{}, i.e. \hypx{}AFormer-S18, obtains equivalent performance to CAFormer-S18.
The global context in earlier layers does not affect categorization performance.
This aligns with our observation that \hypx{} learns local features in earlier stages (see Section~\ref{sec:analysis}).

The model assembles achieve competitive performance without additional training.
We find that \hypx{}Former-S18 and ConvFormer-S18 differ in about $50\%$ of the wrongly classified images.
With a simple ensemble of these two models by mean pooling the predictions, namely \hypx{} / Conv, the accuracy improves to $84.0\%$.
By adding H$_\text{b}$Former-S18 and CAFormer-S18 to the ensemble, i.e. \hypx{} / Conv / H$_{\text{b}}$ / CA, the accuracy further increases to $84.7\%$.

Tab.~\ref{tab:imagenet_hs} reports results for higher-resolution inputs.
Fine-tuning on a resolution of $384\,\text{px}^2$ puts \hypx{}Former-S18 with an accuracy of $84.7\%$ ahead of ConvFormer-S18.
By fine-tuning the unmodified \hypx{}Former-S18 $512\,\text{px}^2$ the accuracy is slightly increased to $84.8\%$.
MaxViT-T trained on equal resolution performs significantly better while requiring more MACs.
 
First of all, our results support the assumptions on the MetaFormer\,\cite{yu2022metaformer_baselines} as a strong baseline model and the expressiveness of Hyena.
Interestingly, we observe that features produced by different token mixers can be incompatible.
Moreover, we close the gap between ConvNets and Transformers with a radical new approach: A ConvNet for vision without a predefined neighborhood.

\subsection{Ablation Study}

We test different aspects of \hypx{}Former-S12.
The training is conducted on IN-1k and mainly follows the procedure described in Section~\ref{sec:eval:image_classifciation}, but, if not otherwise stated, we reduce the number of epochs from 310 to 160 and adjust the cosine decay accordingly if not otherwise stated. 
Tab.~\ref{tab:ablation} reports the results of the ablation study.

\begin{table}[t]
\caption{
Effect of different ablations on the IN-1k top-1 accuracy.
}
\label{tab:ablation}
\centering
\addtolength{\tabcolsep}{-3.pt}
\begin{tabular}{lr}
\toprule[1pt]
 & Top-1\\
\midrule[1pt]
\hypx{}Former-S12 (Baseline) & 80.3 \\
\midrule[1pt]
\multicolumn{2}{l}{Kernel Size for Global Convolution}\\
\midrule[1pt]
$ \left[55^2, 27^2, 13^2, 7^2\right] $ & 80.3  \\
$ \left[27^2, 13^2, 7^2, 3^2\right] $ & 80.1  \\
$ \left[9^2, 9^2, 9^2, 9^2\right] $ & 80.3  \\
\midrule[1pt]
\multicolumn{2}{l}{Token Mixer}\\
\midrule[1pt]
Hyena & 79.9 \\
H$_{\text{b}}$ & 81.0 \\
\hypx{} with Separable Conv. & 79.9 \\
\midrule[1pt]
\multicolumn{2}{l}{LayerNorm (LN)} \\
\midrule[1pt]
\hypx{}Former-S18 & 83.2 \\
\hypx{}Former-S18 without LN & 83.0 \\
\midrule[1pt]
\multicolumn{2}{l}{Network Depth} \\
\midrule[1pt]
\hypx{}Former-S4 & 73.4 \\
ConvFormer-S4 & 73.0 \\
\bottomrule[1pt]
\end{tabular}
\end{table}

\paragraph{Kernel size.}
The global convolution is the main component of \hypx{}Former and is almost twice as large as the feature map, such that each output position can ``see'' all input positions, similar to attention.
Halving the kernel size has no effect on performance while applying only a quarter of the original kernel size causes a slight drop in accuracy of $0.2\%$.
Interestingly, using a constant kernel size of 9 causes no accuracy drop.
The hierarchical structure of the network counteracts the loss of global context in each layer.
However, once the layers in the later stages lose feature map coverage the accuracy is negatively impacted.
Overall, \hypx{}Former has an inherent robustness to changes in hyperparameters.

\paragraph{Other token mixers.}
We already compared the runtime of different token mixers (see Fig.~\ref{fig:runtime}).
The capabilities of token mixers can also vary drastically even within the same architecture~\cite{yu2022metaformer_baselines}.
Bidirectional instead of causal sequence modeling with Hyena (H$_\text{b}$) significantly boosts the top-1 accuracy from $79.9\%$ to $81.0\%$.
Adding a fixed neighborhood definition (\hypx{}) decreases the accuracy by $0.7\%$.
A reason for this decrease could be that the image border is more prominent in \hypx{} due to 3$\times$ more zero values in the input.
We observed that \hypx{} prefers solutions focusing on the horizontal and vertical direction while H$_\text{b}$ shows more complex kernels (see Fig.~\ref{fig:filters}).
To test our hypothesis, we investigate spatially separable convolutions focusing on the main axes and observe a further drop in accuracy by $0.4\%$.
Interestingly, this restriction has a similar effect as the original causal Hyena.
We hypothesize that more complex positional embeddings not preferring a particular direction could improve \hypx{}.
However, this remains for future work.

\paragraph{Normalization for stability.}
By adding a layer normalization~\cite{ba2016layer} after the multiplication of query $q$ and value $k$ (c.f. Fig.~\ref{fig:hyena_pixel_operator}), the accuracy improves slightly by $0.2\%$ with the regular training setting.
Next to the minor improvement, the normalization stabilizes the training of larger network variants, i.e., \hypx{}Former-B36. 

\paragraph{Network depth and context size.}
While ConvNets typically require many layers to view the complete input image, \hypx{} ideally only needs one layer.
We investigate whether we can reduce network depth while increasing a layer's context size by creating two shallow networks:
\hypx{}Former-S4 and ConvFormer-S4 with one block per stage.
The accuracy on IN-1k differs by $0.4\%$ in favor of \hypx{}Former-S4.
While this supports our hypothesis, building a large ERF with a hierarchical structure and small kernels is also effective because of its multiplicative effect on the receptive field~\cite{luo2016understanding}.

\subsection{Downstream Tasks}

\begin{figure}[t]
\centering
\includegraphics[width=0.95\linewidth]{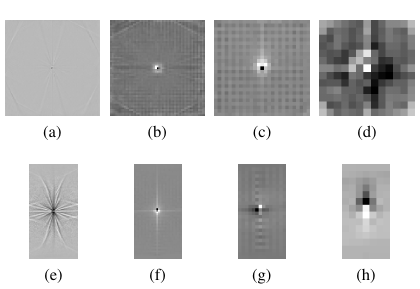}
\vspace{-.5em}
\caption{
Hand-picked normalized mean kernel weights from each stage of the 2D global convolution layers in \hypx{}Former-S18 (a)\,-\,(d) and the reshaped 1D global convolution layers in H$_\text{b}$Former-S18 (e)\,-\,(h). 
Note that for H$_\text{b}$ we wrap the kernel for a specific location.
The kernel would wrap differently at other evaluation positions due to the nature of the 1D convolution and the flattened input image patches (see Fig.~\ref{fig:preview}).
}
\label{fig:filters}
\end{figure}

\paragraph{Object detection and instance segmentation on MS~COCO.}
Following common practice~\cite{liu2021swin, liu2022convnet}, we evaluate the localization properties of \hypx{}Former-S18 with the Cascade Mask R-CNN~\cite{cai2018cascade} on MSCOCO~\cite{lin2014coco}.
The \hypx{} feature maps are extracted at each stage and passed through an additional stage-specific layer normalization. 
With the MMDetection framework~\cite{mmdetection}, we train the model with AdamW, a batch size of 16, a learning rate of $2e^{-5}$, and a stochastic depth of 0.4 for a 3$\times$ schedule (36 epochs), halving the learning rate after 27 and 33 epochs.
Moreover, we apply multi-scale training, i.e., resizing of the shorter side between 480 and 800 pixels and limiting the longer side to 1333 pixels.
Tab.~\ref{tab:coco} reports the results.
The reference models also investigate the downstream performance of a given backbone using the same framework and share a similar computational complexity.
\hypx{}Former-S18 achieves the best performance in object detection with a precision of $52.6\,\text{AP}^{\text{b}}$, outperforming CSWin-T~\cite{dong2022cswin} by $0.1\,\text{AP}^{\text{b}}$, CAFormer-S18~\cite{yu2022metaformer_baselines} by $0.3\,\text{AP}^{\text{b}}$ and ConvNeXt-T~\cite{liu2022convnet} by $2.2\,\text{AP}^{\text{b}}$.
A similar situation can be observed for instance segmentation with a precision of $45.6\,\text{AP}^{\text{m}}$.
CSwin-T, CAFormer-S18, and ConvNeXt-T are trailing by $0.3\,\text{AP}^{\text{m}}$, $0.4\,\text{AP}^{\text{m}}$, and $1.9\,\text{AP}^{\text{m}}$, respectively.
For both tasks, the superior performance can be attributed to the better localization capabilities with higher $AP_{75}$, while $AP_{50}$ is comparable or slightly lower than for the competition.
One reason for the improved localization could be that the image borders are present for each \hypx{} layer at every pixel position and serve as reference guides (see Fig.~\ref{fig:filters}).
Furthermore, large filters enable the model to better recognize object shapes being more similar to human vision~\cite{ding2022scaling}.

\begin{table}[t]
\caption{Object detection and instance segmentation results on the MS~COCO validation set with Cascade Mask R-CNN. Input resolution is $800\times1333$ (except MaxViT with $896\times896$).}
\label{tab:coco}
\centering
\addtolength{\tabcolsep}{-3.5pt}
\begin{tabular}{lrrrrrrrr}
\toprule[1pt]
Model & \#Param. & $\text{AP}^{\text{b}}$ & $\text{AP}^{\text{b}}_{50}$ & $\text{AP}^{\text{b}}_{75}$ & $\text{AP}^{\text{m}}$ & $\text{AP}^{\text{m}}_{50}$ & $\text{AP}^{\text{m}}_{75}$ \\
\midrule[1pt]
Swin-T~\cite{liu2021swin} & 86M  & 50.4 & 69.2 & 54.7  & 43.7 & 66.6 & 47.3 \\
ConvNeXt-T~\cite{liu2022convnet}  & 86M  & 50.4 & 69.1 & 54.8 & 43.7 & 66.5 & 47.3 \\
SLaK-T~\cite{liu2023more} &  -  & 51.3 & 70.0 & 55.7   & 44.3 & 67.2 & 48.1 \\
ConvFormer-S18~\cite{yu2022metaformer_baselines} & -  & 51.5 & 70.7 & 55.8   & 44.6 & 67.8 & 48.2 \\
MaxViT-T~\cite{tu2022maxvit} & 69M  & 52.1 & 71.9 & 56.8 & 44.6 & 69.1 & 48.4 \\
CAFormer-S18~\cite{yu2022metaformer_baselines} & -  & 52.3 & 71.3 & 56.9   & 45.2 & 68.6 & 48.8 \\
CSWin-T~\cite{dong2022cswin} & 80M & 52.5 & 71.5 & 57.1  & 45.3 & 68.8 & 48.9 \\

\rowcolor{gray!25!}
\hypx{}Former-S18  & 84M & 52.6 & 71.3 & 57.3 & 45.6 & 68.7 & 49.5 \\

\bottomrule[1pt]
\end{tabular}
\end{table}

\begin{table}[t]
\caption{Semantic segmentation on ADE20k validation set using UperNet~\cite{xiao2018unified} with an input resolution of $512^2$. MACs are calculated based on an input resolution of $512\times2048$.}
\label{tab:ade20k}
\centering
\addtolength{\tabcolsep}{-3.5pt}
\begin{tabular}{lrrrr}
\toprule[1pt]
   Model & \#Param. & MACs & mIoU & mIoU$_{\text{MS}}$  \\
\midrule[1pt]
Swin-T~\cite{liu2021swin} & 60M & 945G & 44.5 & 45.8 \\
ConvNeXt-T~\cite{liu2022convnet} & 60M & 939G & 46.0 & 46.7 \\
SLaK-T~\cite{liu2023more} & 65M & 936G & 47.6 & - \\
InternImage-T~\cite{wang2023internimage} & 59M & 944G & 47.9 & 48.1 \\
ConvFormer-S18~\cite{yu2022metaformer_baselines} & 54M & 925G & - & 48.6 \\
\rowcolor{gray!20!}
\hypx{}Former-S18 & 56M & 928G & 48.1 & 48.7 \\
CAFormer-S18~\cite{yu2022metaformer_baselines} & 54M & 1024G & - & 48.9 \\
CSWin-T~\cite{dong2022cswin} & 60M & 959G & 49.3 & 50.7 \\
BiFormer-S~\cite{zhu2023biformer} & - & - & 49.8 & 50.8 \\
\bottomrule[1pt]
\end{tabular}
\end{table}

\paragraph{Semantic segmentation on ADE20k.} 
We evaluate the downstream performance on semantic segmentation with UperNet~\cite{xiao2018unified} on the ADE20k benchmark~\cite{zhou2019semantic}, following related work~\cite{liu2021swin}.
We base our implementation on MMSegmentation~\cite{mmsegmentation} and train with AdamW for 160k steps with a batch size of 16, a learning rate of $1e^{-4}$, and a stochastic depth of 0.4.
Tab.~\ref{tab:ade20k} reports the results.
\hypx{}Former-S18 beats Swin-T by $4.2$\,mIoU, ConvNeXt-T by $2.1$\,mIoU, SLaK-T by $0.5$\,mIoU and InternImage-T by $0.2$\,mIoU in the single scale setting.
CSWin-T and BiFormer-S perform significantly better with an improvement of $1.2$\,mIoU and $1.7$\,mIoU, respectively. 
Semantic segmentation is significantly more difficult for \hypx{}Former-S18 than instance segmentation.
We assume that while global context is relevant, the model has no mechanism to filter the features in a data-driven way similar to attention~\cite{vaswani2017attention} or more sophisticated approaches~\cite{xia2022vision, zhu2023biformer}.
Furthermore, we expect that semantic segmentation will benefit from local texture-focused operations.

\begin{figure}[t]
\centering
\includegraphics[width=0.95\linewidth]{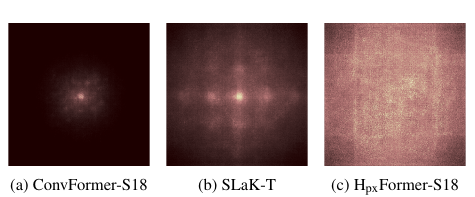}
\caption{
Effective Receptive Field (ERF) of different models sampled over 50 images of size $1024\text{px}^2$ from the IN-1k validation set.
}
\label{fig:erf}
\end{figure}

\section{Analysis}
\label{sec:analysis}

\paragraph{Effective receptive field.}
The Effective Receptive Field (ERF) measures the influence of each input pixel on the center-most output value by tracking the gradients in a backward pass~\cite{luo2016understanding}.
A large ERF is often associated with a better performance in vision tasks~\cite{ding2022scaling,liu2023more}.
We follow related work~\cite{ding2022scaling,liu2023more} and compare the ERFs~\cite{kim2023dead}.
Fig.~\ref{fig:erf} shows the ERFs of three models.
ConvFormer and SLaK have a strong local bias caused by local convolution as main the building block.
SLaK features off-center areas with high gradients caused by the separable sparse convolution.
\hypx{}Former has a large ERF with no obvious center location, but some vertical and horizontal artifacts.
This finding shows that \hypx{} captures the global image context.

We hypothesize that \hypx{} could benefit from an additional residual connection with a small convolution.
This modification could be particularly helpful for localization and categorization tasks and was already successfully applied for attention-based networks~\cite{tu2022maxvit, dong2022cswin, zhu2023biformer}.
We leave this study for future research.

\begin{figure}[t]
\centering
\includegraphics[width=0.95\linewidth]{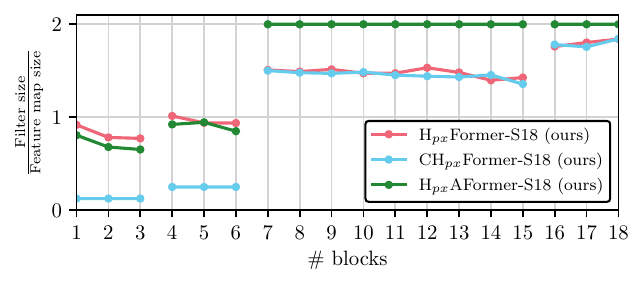}
\vspace{-1em}
\caption{
Learned filter sizes in \hypx{}Former-S18 relative to feature map sizes at different network depths.
For attention, we set the relative feature map coverage to 2, and for convolution, we use the kernel size relative to the feature map size.
Note that the feature map coverage can be greater than one because the kernel size of \hypx{} is almost twice the feature map size.
}
\label{fig:filter_scale}
\end{figure}

\begin{figure}[t]
\centering
\includegraphics[width=0.95\linewidth]{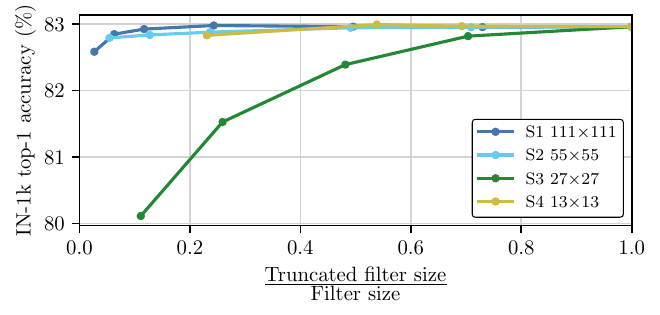}
\vspace{-1em}
\caption{
Impact of truncated filters in \hypx{}Former-S18 on the top-1 IN-1k accuracy.
For each stage (S), we modify the large kernels within the current stage by setting all values to zero that are larger than the relative filter size.
}
\label{fig:truncation}
\end{figure}

\paragraph{Truncate kernel in trained models.}
Due to the learnable decay parameter in \hypx{}, we can estimate the required kernel size at different depths.
By setting all values of $\text{Window}(t_x, t_y)$ to zero that are smaller than 0.05, we can measure the diameter of the non-zero values.
Fig.~\ref{fig:filter_scale} shows the mean relative feature map coverage of the token mixers in each block.
\hypx{} learns similar kernel sizes at the same stage regardless of other token mixers involved in earlier or later stages.
The coverage in each stage stays almost constant, while former layers of a stage have slightly larger kernels.
Overall, the optimal feature map coverage increases with depth, consistent with the observation of \citet{romero2021flexconv}.
To further investigate the importance of filter size, we truncate the filters within each stage of a pre-trained \hypx{}Former-S18 and visualize the IN-1k classification results in Fig.~\ref{fig:truncation}.
The truncation of the third stage has the biggest impact, with an accuracy drop of almost $-2.9\%$.
Surprisingly, the first and last stage are more local and can benefit from truncation, improving performance slightly.
These insights might help construct better model layouts.

\section{Conclusion}
\label{sec:conclusion}

In this work, we studied whether the Hyena operator is a sufficient replacement for attention in computer vision applications.
We extended Hyena to non-causal, bidirectional sequence modeling and added spatial bias with a fixed pixel neighborhood. 
We found the Hyena formulation useful for training extremely large kernels up to 191$\times$191.
Analyzing trained models with these token mixers showed that bidirectional modeling is sufficient to achieve competitive categorization accuracy, while a fixed pixel neighborhood hurts the final performance.
However, spatial bias with large kernels improves performance for downstream tasks dependent on exact localization.
Our analysis showed that the ERF for our two-dimensional Hyena lacks the local bias present in other approaches.

In conclusion, our results suggest large, non-causal, bidirectional, spatially unbiased convolution as a promising avenue for future research.



\clearpage

\begin{ack}
This research has been funded by the Federal Ministry of Education and Research of Germany under grant no. 01IS22094C WEST-AI.
\end{ack}



\clearpage
\appendix

\section{Overview}

In this supplementary material to the paper ``HyenaPixel: Global Image Context with Convolutions'', we investigate the properties and learned weights of our non-causal Hyena (H$_\text{b}$) and HyenaPixel (\hypx{}) operators.
We extend our investigation of the Effective Receptive Field (ERF) in Sec.~\ref{sec:erf} for different models.
Furthermore, we provide visual explanations for models with small and large filters to study the important pixels for categorization (Sec.~\ref{sec:expl}).
We look at the learned kernels of H$_\text{b}$ and \hypx{} to gain insight into the effect of large filter sizes and spatial bias on the filter structure (Sec.~\ref{sec:filt}).
Finally, we extend our ablation study (Sec.~\ref{sec:ext_ablation_study}).

\begin{figure}[t]
\centering
\includegraphics[width=0.95\linewidth]{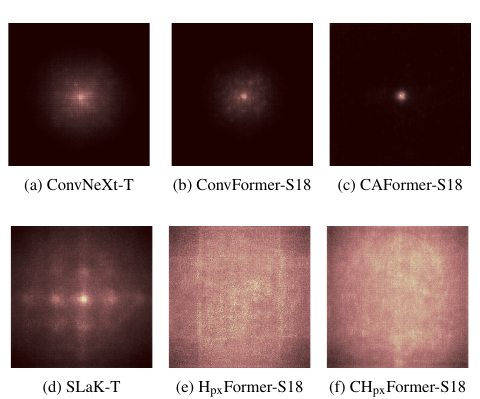}
\caption{
Effective Receptive Field (ERF) of different models sampled over 50 images of size $1024\text{px}^2$ drawn from the IN-1k validation set.
}
\label{fig:erf2}
\end{figure}

\section{Effective Receptive Field}
\label{sec:erf}

The Effective Receptive Field (ERF) measures the influence of each input pixel on the centermost output value by tracking the gradients in a backward pass~\cite{luo2016understanding}.
A large ERF is often associated with a better performance in vision tasks~\cite{ding2022scaling,liu2023more}.
We follow related work~\cite{ding2022scaling,liu2023more} and compare the ERF~\cite{kim2023dead} by sampling 50 images from the ImageNet-1k (IN-1k) validation set with a resolution of $1024 \,\text{px}^2$.

Fig.~\ref{fig:erf} compares the ERFs of six models.
ConvNeXt~\cite{liu2022convnet} and ConvFormer~\cite{yu2022metaformer_baselines} share similar bell-shaped fields with a local bias, with the ERF of ConvNeXt being slightly larger.
Applying attention~\cite{vaswani2017attention} layers in the last two stages (i.e., CAFormer~\cite{yu2022metaformer_baselines}) increases the local focus.
However, the model also interacts with more distant image regions to a smaller extent.
Next to local bias, SLaK~\cite{liu2023more} features off-center areas with high gradients caused by the separable sparse convolution.
In contrast, \hypx{}Former has no areas of high gradient or center location.
Instead, the ERF covers the entire input image with a slight drop-off at the edges.
Small convolutions in the first two stages (i.e., C\hypx{}Former) still do not add local bias but smooth the ERF.

\begin{figure}[t]
\centering
\includegraphics[width=0.95\linewidth]{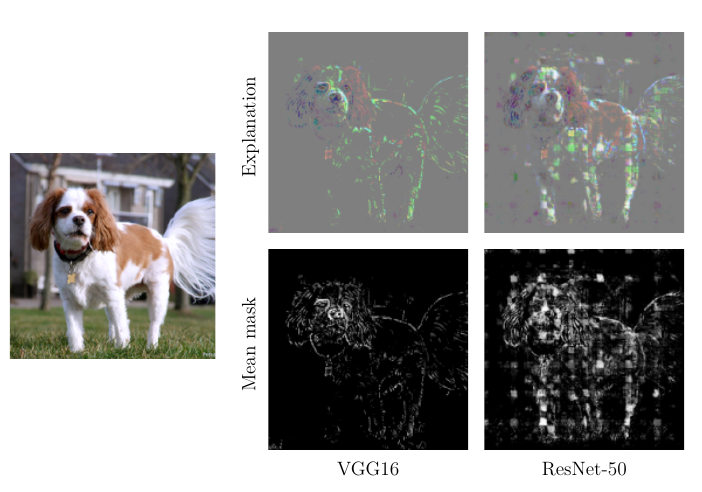}
\caption{
Fine-grained visual explanations (right) of an input image (left) generated with the FGVis method~\protect\cite{wagner2019interpretable}.
The explanation is calculated by multiplying the image $x$ with the inverse mask $1-m$.
We add a grayscale image multiplied by $m$ for better visibility.
The mean mask is the mean value along the color channel of $m$.
The mask values are raised by exponentiation with 7 for better visibility.
The input image is from the IN-1k validation set.
}
\label{fig:explanations_sample}
\end{figure}

\begin{figure*}[p]
\centering
\includegraphics[width=0.95\linewidth]{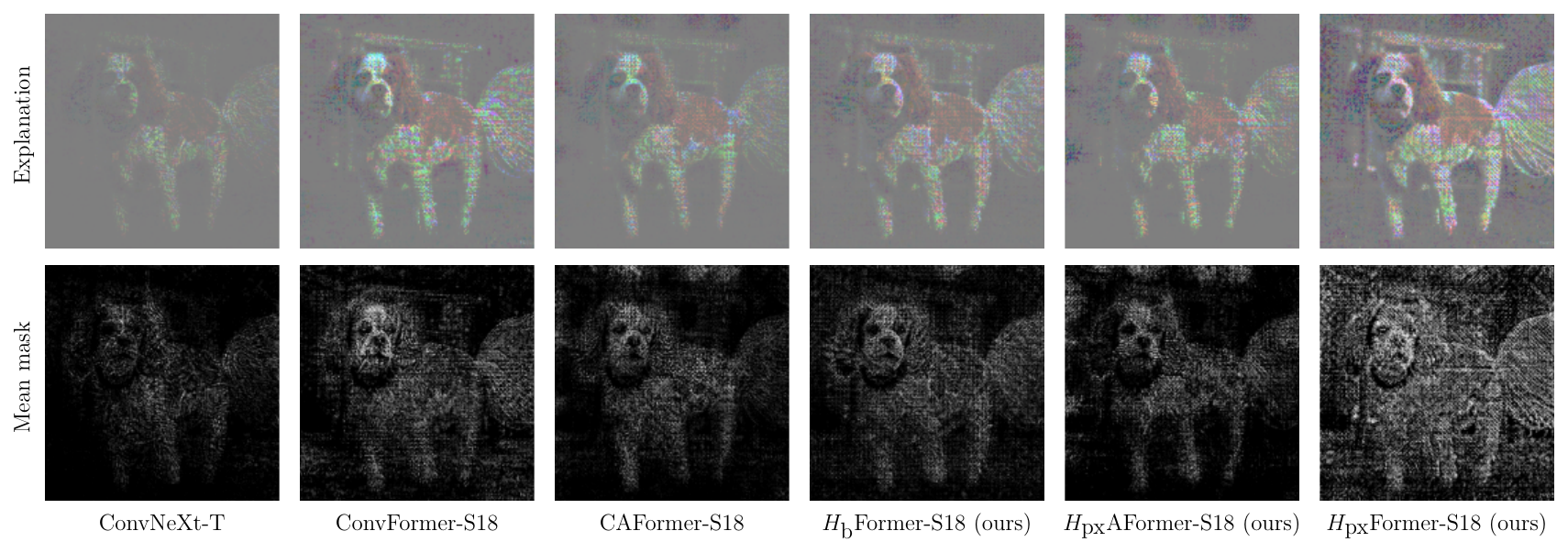}
\vspace{-0.3cm}
\caption{
Fine-grained visual explanations for different models generated with the FGVis method~\protect\cite{wagner2019interpretable}.
Fig.~\ref{fig:explanations_sample} shows and explains the input image and the visualization strategy.
}
\label{fig:explanations}
\end{figure*}

\begin{figure*}[p]
\centering
\includegraphics[width=.95\linewidth]{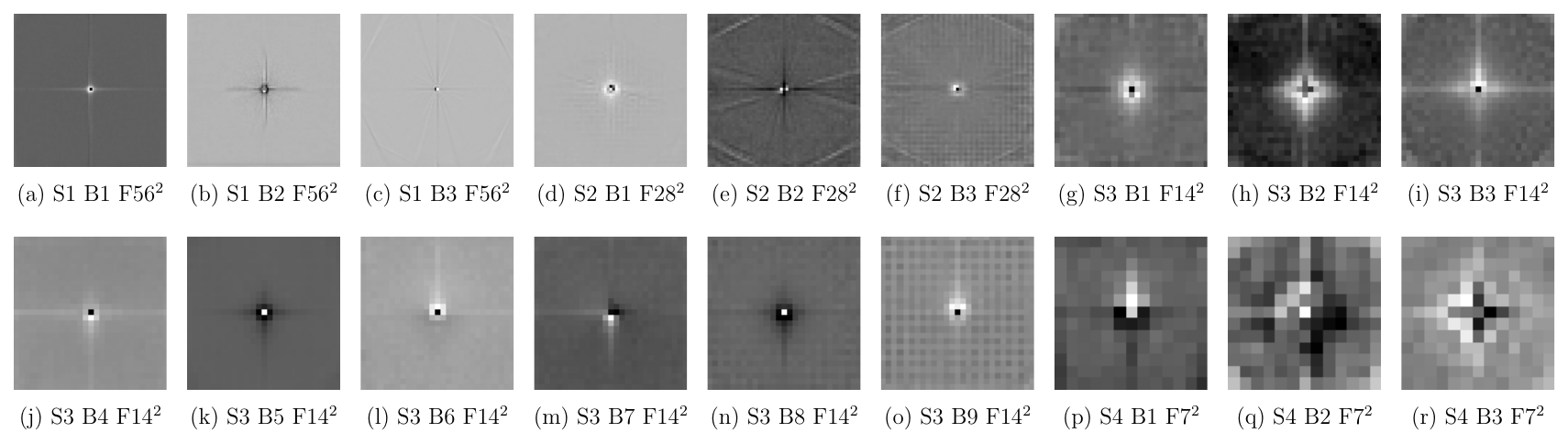}
\vspace{-0.3cm}
\caption{
Normalized mean kernel weights for the 2D global convolution layers in the H$_\text{px}$Former-S18 by stage (S), block (B), and feature map size $F$.
The width and height of the kernel are given by $2F -1$, which provides the kernel with almost four times as many elements as the feature map.
}
\label{fig:mean_kernels_hypx}
\end{figure*}

\begin{figure*}[p]
\centering
\includegraphics[width=.95\linewidth]{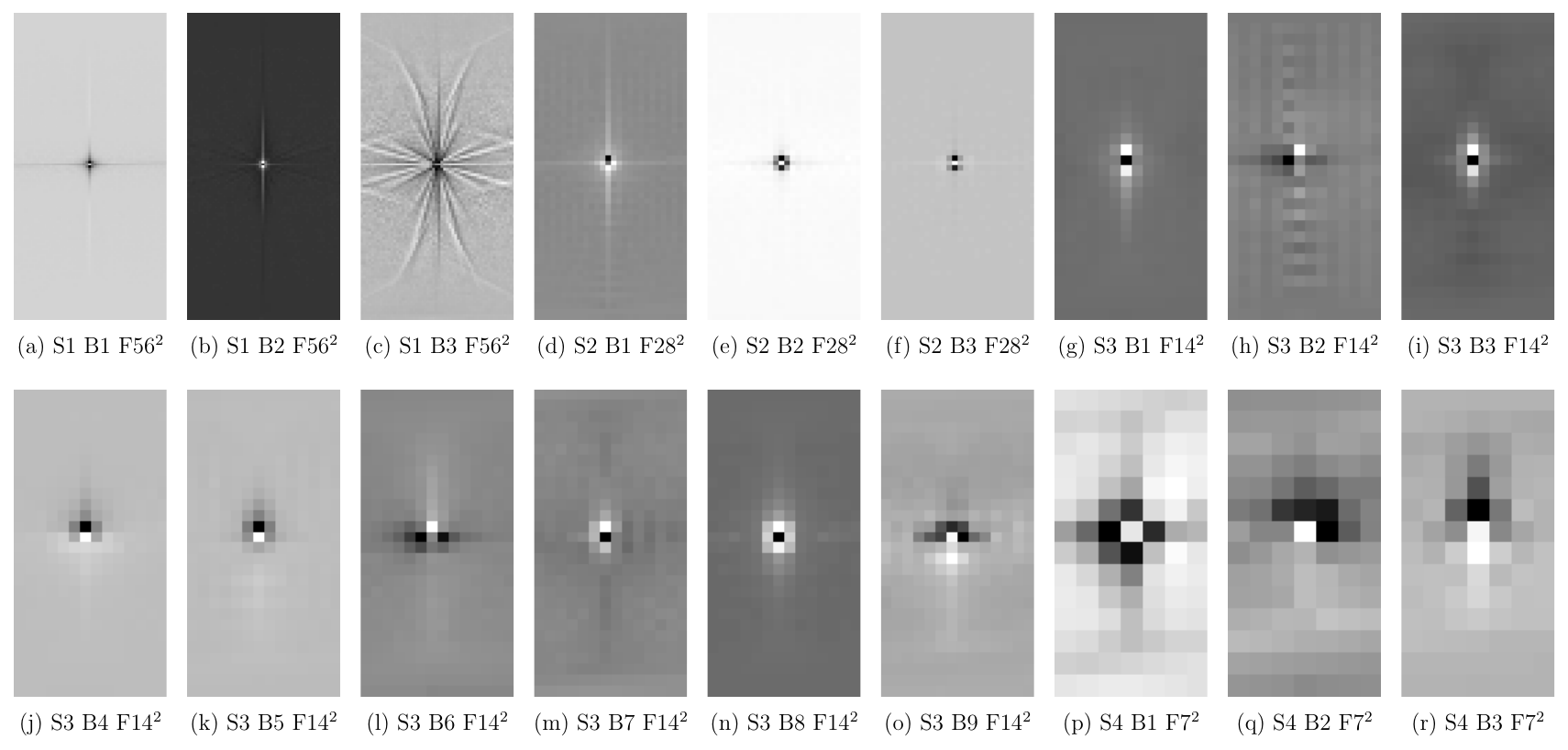}
\vspace{-0.3cm}
\caption{
Normalized and reshaped mean kernel weights for the 1D global convolution layers in the H$_\text{b}$Former-S18 by stage (S), block (B), and feature map size $F$.
The kernel size is defined by $2F^2 -1$.
The visualization is not square because the 1D filter has $F^2 -1$ more elements than the feature map, and we reshape the filter with a fixed width of $F$ to account for the wrapping.
Note that the visualization only shows the 2D reconstruction for the centermost pixel.
The kernel would wrap differently at other evaluation positions due to the nature of the 1D convolution and the flattened input image.
}
\label{fig:mean_kernels_hyena}
\end{figure*}

\section{Fine-grained Visual Explanations}
\label{sec:expl}
Visual explanations can be formulated as an optimization problem: Find a mask that selects all pixels relevant to the object class with the highest probability.
When this mask is subtracted from the original image, the highest class probability will switch to a different category.
One issue with this approach is that it can also produce or favor adversarial masks.
We utilize the adversarial defense and visualization method FGVis~\cite{wagner2019interpretable} by masking gradients that would move the output of activation layers beyond an upper or lower bound.
The bounds are determined by measuring the outputs of these layers with the unmasked image of interest.
We follow the training strategy proposed by \citet{wagner2019interpretable}.
The loss function is given by
\begin{equation}
    y_e - \lambda \left\Vert m \right\Vert_1 \text{,}
\end{equation}
where $y_e$ is the softmax score of the target class for the evidence $e = x \cdot m$ of the input image $x$ and the mask $m$.
$\lambda$ is the weight of the sparsity term.
We optimize for 500 iterations with SGD and a learning rate of $0.1$.
The weight mask is initialized to 1.
The training is stopped early if the class with the highest softmax score changes.
The value of $\lambda$ is determined by decreasing its value, starting at $\lambda= 1e^{-4}$, until the first early stopping criterion is met.
We add further regularization by normalizing the gradient, clamping the mask values between 0 and 1, and adding noise to the input.

Fig.~\ref{fig:explanations_sample} shows explanations for VGG~\cite{simonyan2014very} and ResNet~\cite{he2016deep}.
VGG has a clear focus on the dog's head and edge information.
ResNet also incorporates texture information and other body parts for categorization.
In addition, we also find the checkerboard pattern.
This pattern is attributed to architectural details~\cite{wagner2019interpretable}.

Fig.~\ref{fig:explanations} shows explanations for transformer-based~\cite{vaswani2017attention} architectures.
The resulting explanations are smoother and more dense than those produced by classical convolutional neural networks (ConvNets), c.f. Fig.~\ref{fig:explanations_sample}.
For all models, there is some influence by the background, while Conv\-NeXt has the lowest and \hypx{}\-Former has the highest background dependency.
The models focus on texture and the dog's facial features. 
\hypx{}\-Former and Conv\-Former are also dependent on edge information. 
This dependency disappears when the last two stages are filled with attention layers instead (CAFormer and \hypx{}AFormer)  and even adds an offset to the inside relative to the border.
Interestingly, networks with attention also focus on regions with low information density, such as the house's roof.
This phenomenon is most likely related to register tokens~\cite{darcet2023vision}.

\section{Learned Convolution Kernels}
\label{sec:filt}

We visualize the learned global convolution kernels of \hypx{}Former-S18 with spatial bias and H$_{\text{b}}$ without spatial bias.
Fig.~\ref{fig:mean_kernels_hypx} shows the kernels of \hypx{}.
The horizontal and vertical lines are the most prominent features.
Earlier stages show a clear center focus, while later stages can have high magnitude off-center elements. 
Furthermore, some filters feature grid patterns or even patterns that could be described as snowflake-shaped.
Similar structures can be observed in the filters learned by H$_{\text{b}}$Former depicted in Fig.~\ref{fig:mean_kernels_hyena}.
However, H$_{\text{b}}$ filters appear smoother and better centered. 
We hypothesize that the sequential modeling increases the robustness against edge effects. 
Also, this naturally increases the complexity of the filter, as the 1D filters wrap around the image edges, effectively creating a unique filter for each position.

\section{Extension of the ablation study}
\label{sec:ext_ablation_study}

\paragraph{Attention pooling with register tokens.}
Attention pooling selects relevant tokens within a sequence based on learned queries~\cite{touvron2021augmenting}.
Intuitively, this helps the network to focus on certain parts of the feature map, like foreground pixels.
We extend attention pooling by register tokens~\cite{darcet2023vision} to act as an ``attention fallback'' if the image contents are irrelevant to the query token.
Mean pooling outperforms attention pooling by $0.1\%$.

\end{document}